\documentclass[runningheads]{llncs}

\usepackage{eccv}

\usepackage{eccvabbrv}

\usepackage{graphicx}
\usepackage{booktabs}

\usepackage[accsupp]{axessibility}  

\usepackage{hyperref}

\usepackage{orcidlink}
\usepackage{amsmath}
\usepackage{mathrsfs}
\usepackage{amsfonts}
\usepackage{amssymb}
\usepackage{calrsfs}
\usepackage{tabularray}
\usepackage{multirow}
\usepackage{wrapfig}
\usepackage{xcolor}
\definecolor{mygreen}{RGB}{152,160,93}
\definecolor{myorange}{RGB}{208,104,79}

\begin{document}

\title{
InfoNorm: Mutual Information Shaping of Normals for Sparse-View Reconstruction
} 

\titlerunning{InfoNorm}

\author{
Xulong Wang$^{1,2*}$ \quad
Siyan Dong$^{3*}$ \quad
Youyi Zheng$^{1\dagger}$ \quad
Yanchao Yang$^{3,4\dagger}$
}

\authorrunning{X.~Wang \& S.~Dong et al.}


\institute{
$^{1}$ State Key Lab of CAD\&CG, Zhejiang University $^{2}$ Chohotech co. ltd. \\
$^{3}$ Institute of Data Science, The University of Hong Kong \\ 
$^{4}$ Department of Electrical and Electronic Engineering, The University of Hong Kong
}

\maketitle

\def \thefootnote{*}\footnotetext{Equal contributions (wangxulong@zju.edu.cn, siyan3d@hku.hk).}
\def \thefootnote{$\dagger$}\footnotetext{Corresponding authors.}
\def \thefootnote{2}\footnotetext{This work was done during the author's internship at Chohotech Co. ltd..}

\begin{abstract}

3D surface reconstruction from multi-view images is essential for scene understanding and interaction. However, complex indoor scenes pose challenges such as ambiguity due to limited observations. Recent implicit surface representations, such as Neural Radiance Fields (NeRFs) and signed distance functions (SDFs), employ various geometric priors to resolve the lack of observed information. Nevertheless, their performance heavily depends on the quality of the pre-trained geometry estimation models. To ease such dependence, we propose regularizing the geometric modeling by explicitly encouraging the mutual information among surface normals of highly correlated scene points. In this way, the geometry learning process is modulated by the second-order correlations from noisy (first-order) geometric priors, thus eliminating the bias due to poor generalization. Additionally, we introduce a simple yet effective scheme that utilizes semantic and geometric features to identify correlated points, enhancing their mutual information accordingly. The proposed technique can serve as a plugin for SDF-based neural surface representations. Our experiments demonstrate the effectiveness of the proposed in improving the surface reconstruction quality of major states of the arts.
Our code is available at: 
\url{https://github.com/Muliphein/InfoNorm}.

\end{abstract}
\section{Introduction}

\begin{figure*}[!t]
\centering
\includegraphics[width=1.0\linewidth]{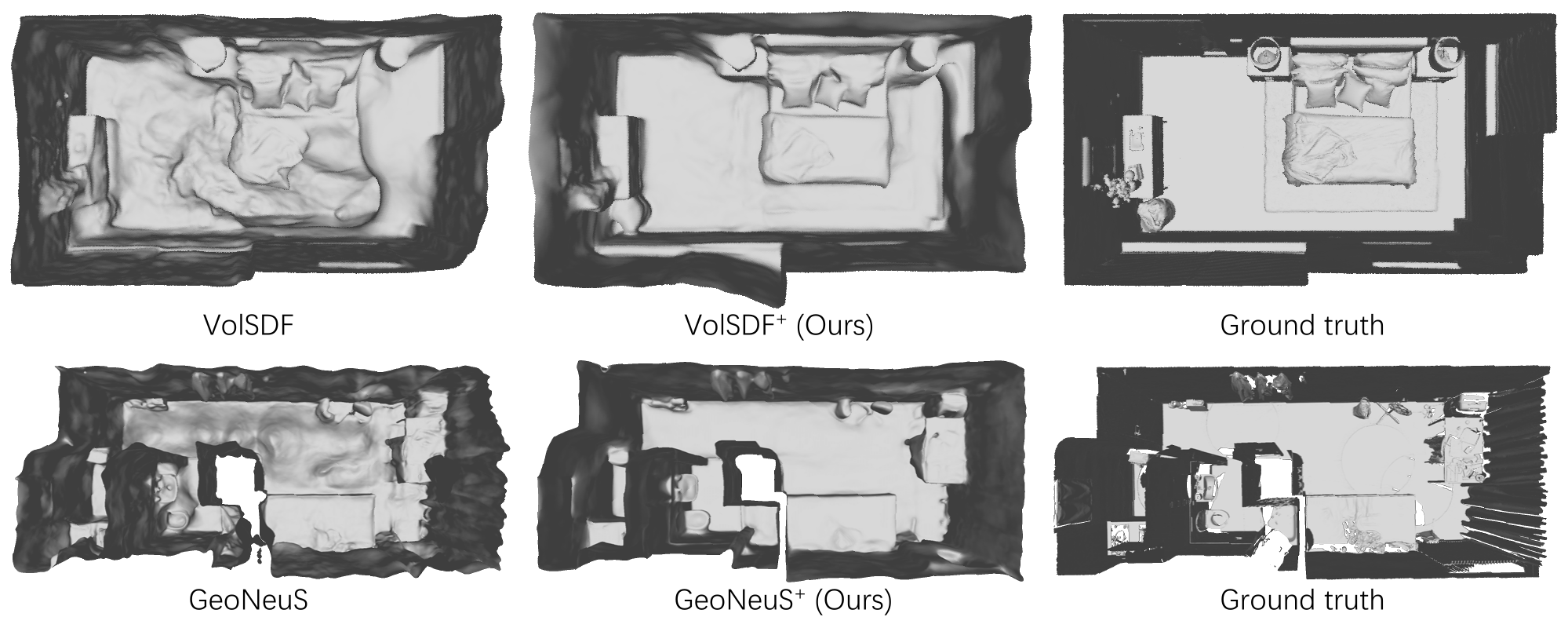}
\caption{
3D scene reconstruction from sparse views on Replica~\cite{straub2019replica} (first row) and ScanNet++~\cite{yeshwanthliu2023scannetpp} (second row). 
By enforcing the mutual information between the normals of highly correlated scene points, the proposed method can effectively enhance the reconstruction quality of the baselines (VolSDF~\cite{yariv2021volume} and GeoNeuS~\cite{fu2022geo}).
}
\label{fig:teaser}
\end{figure*}

3D surface reconstruction from multi-view images is an essential task in the computer vision and graphics community, with practical applications including content creation for virtual reality and robot-scene interaction. 
However, indoor scenes present challenges due to the large and complex scenes captured from sparse viewpoints. These challenges encompass issues such as occlusion and ambiguity arising from limited observations. As a result, traditional Multi-View Stereo (MVS) methods~\cite{yao2018mvsnet,wang2021patchmatchnet}, which require substantial overlap among images, may not produce satisfactory results in these scenarios.

Recently, implicit scene representations, e.g., Neural Radiance Fields (NeRFs)
\cite{mildenhall2020nerf}, 
have been proposed to encode 3D scenes into a set of neural network parameters, with subsequent works enhancing the representational capacity for geometry to enable surface extraction. 
The key idea involves incorporating signed distance functions (SDFs)~\cite{park2019deepsdf} into the learned radiance field.
Although NeRFs with SDF produce high-quality reconstructions of simple scenes with sufficient images, they still underperform in large and complex scenes captured from sparse viewpoints. 
More recent approaches, such as NeuRIS~\cite{wang2022neuris} and MonoSDF~\cite{yu2022monosdf}, propose using monocular geometric priors to guide the learning of the radiance field. 
However, they can be affected by inevitable errors and noise in the monocular geometry estimation modules.

\def \thefootnote{1}\footnotetext{We use the term ``shaping'' to refer to the process of optimizing network parameters to enforce consistencies in the scene geometry via aligning the gradients.}

We explore geometric regularization in a second-order perspective. 
We argue that both the geometry itself and the correlation of geometry among scene regions are crucial for surface reconstruction, 
especially when dealing with varying indoor scenes under sparse views with noisy geometry estimates. 
The correlation is expressed as mutual information between scene entities under random perturbations of the network weights. 
More explicitly, 
we propose to enforce the mutual information between the normals of two points in the scene that are considered geometrically correlated,
where the mutual information can be efficiently computed by measuring the cosine similarity between the gradients of the normals with respect to the perturbed weights.
Further, we leverage a combination of pre-trained semantic and geometric features \cite{do2020surface,caron2021emerging,kirillov2023segment} to identify the correlated scene regions that endorse high mutual information. 
This results in an easy-to-use geometric shaping$^1$ technique that can be applied to any SDF-based neural radiance fields. 
To thoroughly evaluate its effectiveness, 
we test the proposed with multiple state-of-the-art neural surface reconstruction models. 
The experimental results confirm the effectiveness of the proposed shaping technique, showing improved surface reconstruction quality (e.g., Fig.~\ref{fig:teaser}) on challenging scenes with limited views.
To summarize:
\begin{itemize}

\item 
We propose to leverage second-order geometric correlation in the form of mutual information between normals to regularize the surface reconstruction of SDF-based NeRF representations.

\item 
We develop a pipeline that efficiently enforces the mutual information of the normals of points, deemed geometrically correlated with pre-trained multimodal features, to improve the reconstruction quality of the scene.

\item 
We verify the effectiveness of the proposed second-order regularization technique across a broad spectrum of baselines and demonstrate its usefulness as an easy-to-use plugin in improving 3D surface modeling.
\end{itemize}

\section{Related Work}

\paragraph{3D Reconstruction with neural radiance fields (NeRFs).}
NeRF~\cite{mildenhall2020nerf} is originally designed for novel view synthesis,
with its variants serving as scene representations~\cite{liu2023robust,xu2023grid,yan2023nerf,yu2023dylin,xiangli2022bungeenerf} for tasks related to editing~\cite{bao2023sine,zhu2023i2sdf,weder2023removing}, semantic segmentation~\cite{xu2023jacobinerf,liu2023semantic,liu2023unsupervised,chen2022sem2nerf}, 3D shape generation~\cite{tertikas2023partnerf,metzer2023latent}, and so on.
We mainly focus on neural surface representations that enable the extraction of high-quality 3D meshes from the trained NeRFs. 

Early works such as NeuS~\cite{wang2021neus} and VolSDF~\cite{yariv2021volume} combine signed distance functions (SDFs) with volume rendering to extract high-quality surfaces. 
Recently, SparseNeuS~\cite{long2022sparseneus} introduces a cascaded geometry reasoning framework that generalizes well to novel scenes. 
GeoNeuS~\cite{fu2022geo} uses surface points from structure-from-motion and neighbor-view patches to reconstruct surfaces, while NeuDA~\cite{cai2023neuda} employs a deformable anchor to adaptively encode geometric details. 
NeAT~\cite{meng2023neat} broadens the scope of reconstruction to arbitrary topology, no longer confining to watertight surfaces. 
Moreover, 
Neuralangelo~\cite{li2023neuralangelo} uses a hash grid for position encoding and implements a coarse-to-fine optimization with numerical gradients. 
Despite their promising results, 
most of these methods need sufficient training views, and their performance declines when dealing with large and complex scenes under sparse viewpoints.

\paragraph{Incorporating priors into NeRFs.}
One can augment scene representations by introducing various priors to NeRFs. 
VDN-NeRF~\cite{zhu2023vdn} normalizes the spatial feature to align with monocular features. 
LERF~\cite{kerr2023lerf} introduces a new branch whose outputs align with DINO~\cite{caron2021emerging} and CLIP~\cite{radford2021learning} features. 
Semantic-NeRF~\cite{zhi2021place} encodes semantic information for segmentation tasks, while JacobiNeRF~\cite{xu2023jacobinerf} shapes the color gradients to encode the mutual information in terms of semantic similarity from DINO~\cite{caron2021emerging} features for label propagation.

For surface reconstruction, 
GeoNeuS~\cite{fu2022geo} incorporates patch similarity from neighboring images and sparse surface points derived from structure-from-motion \cite{schoenberger2016sfm}, while I$^2$-SDF~\cite{zhu2023i2sdf} combines the rendering equation with VolSDF~\cite{yariv2021volume} for better novel views and geometry.  
To cope with complex indoor scenes, 
NeuRIS~\cite{wang2022neuris} incorporates the normal priors as additional geometric supervision. 
Further, MonoSDF~\cite{yu2022monosdf} uses both depth and normal to improve the surface reconstruction quality.
However, these methods highly depend on the quality of the supplemented geometric information. 
In contrast, our method leverages the second-order information in geometric prior, rather than directly using the prior as a first-order supervision signal, which makes it more robust to errors and noise.
There are also methods utilizing strong assumptions derived for human-made scenes. 
For instance, ManhattanSDF~\cite{guo2022neural} works under the assumption that all captured scenes adhere to the Manhattan-world concept, thus, predicting semantic segmentation to identify walls and floors. 
While these methods primarily focus on planar scenes, 
the proposed technique can work for general scenes under noisy correlation information.

\section{Method}

We start with a discussion on the preliminaries in Sec.~\ref{sec:preliminaries} to provide the background on how we can learn from posed images an implicit scene representation (NeRF), improve 3D scene reconstruction via the combination of NeRF and SDF representations, and employ mutual information as a constraint in the NeRF training process. 
Following this, in Sec.~\ref{sec:sdfshaping}, we delve into how mutual information can be used as constraints on NeRF's implied surfaces, so that we can learn better geometries for higher-quality 3D reconstruction of the scene. 
Finally, we elaborate on how we can employ off-the-shelf semantic and geometry features to assist the encoding of mutual information through the proposed geometric shaping framework.

\subsection{Preliminaries} 
\label{sec:preliminaries}

\paragraph{Neural radiance fields.}
NeRF~\cite{mildenhall2020nerf} learns an implicit scene representation from a set of posed images. 
For a specific pixel in an image, 
NeRF samples a list of 3D points, $\{x|x_t=\mathbf{o}+t\mathbf{v}, t\in[t_{near},t_{far}]\}$, from the camera center $\mathbf{o}$ along the viewing direction $\mathbf{v}$ within a bounded space. 
Specifically, 
it employs a Multilayer Perceptron (MLP) to encode the volume density and color of each sampled point. 
We denote all the parameters of NeRF as $\theta$, 
the density function as $\sigma$, and the color function as $c$. 
The pixel's rendered color 
$\hat{\mathbf{C}}$ from a NeRF is calculated by the discrete volume rendering: 
\begin{equation}
\label{equ:volumerendering}
    \begin{split}
    \hat{\mathbf{C}} & (\mathbf{o}, \mathbf{v};\theta) = \sum_{i=1}^{N} \omega(\mathbf{o}, \mathbf{v}, t_i;\theta) c(x_{t_i}, \mathbf{v}; \theta), \\
& \text{ where } \omega(\mathbf{o}, \mathbf{v}, t_i;\theta) = \prod_{j=1}^{i-1}(1-\alpha_j)\alpha_i .
    \end{split}
\end{equation}
Here, $\alpha_i=1-\exp(-\sigma_i\delta_i)$, where $\sigma_i$ is the abbreviation of $\sigma(x_{t_i};\theta)$ and $\delta_i=t_{i+1}-t_i$. 
A photometric loss is utilized to minimize the discrepancy between the rendered color and its ground truth: 
\begin{equation}
\label{eq:photometric_loss}
L_C = ||\mathbf{C} - \hat{\mathbf{C}}||_1,
\end{equation}
so that the parameters $\theta$ can be optimized to fit the scene. 
Despite the ability to handle photorealistic novel view synthesis, 
NeRFs suffer from noisy and unrealistic surface extraction when it comes to 3D reconstruction, as discussed in the literature~\cite{wang2021neus,yariv2021volume}.

\paragraph{NeRFs with signed distance functions.}
To produce high-quality surfaces, 
it is beneficial to use signed distance functions (SDFs)~\cite{curless1996volumetric,park2019deepsdf} to represent geometry.
Consequently, the surface can be reconstructed by extracting the zero-level set~\cite{lorensen1998marching} of the SDF values. 
The combination of NeRF and SDF can be achieved by deriving the NeRF's densities from the SDF, 
or defining Laplace's cumulative distribution function. 
Accordingly, 
$\alpha_i$ in Eq.~\ref{equ:volumerendering} can be represented 
by a function $\Phi$ which converts SDF values to volume densities as 
\begin{equation}
    \alpha_i=\Phi(x, f(x;\theta), i) .
\end{equation}
Generally, NeRFs with an SDF achieve improved surface reconstruction quality, where the key to maintaining the characteristics of SDF is an eikonal loss~\cite{gropp2020implicit}: 
\begin{equation}
\label{eq:eikonal_loss}
L_E = \frac{1}{N}\sum_{i=1}^{N}(||\nabla f(x_i)||_2-1)^2.
\end{equation}

\paragraph{Mutual information in NeRFs.}
\label{sec:mutualinformation}
Mutual information (MI) quantifies the statistical dependence between two random variables, providing an estimate of how much information they share.
It has been introduced for semantic segmentation and motion modeling tasks~\cite{xu2023jacobinerf,zhang2024infogaussian}.  
In this paper, We study the mutual information under perturbations of the NeRF parameters, and focus more on the geometry properties. 
Let $I(p_i)$ and $I(p_j)$ denote two pixels derived from the NeRF rendering process $\mathbb{F}$.
And we assume that a subset of the NeRF weights, 
denoted by $\theta^D$, is perturbed by a random noise vector $\mathbf{n}\in \mathbb{R}^D$ sampled from a uniform distribution on the sphere $\mathbb{S}^{D-1}$ by a small step $\gamma \ll 1.0$. 
The random variables corresponding to the perturbed pixels can be written through a Taylor expansion: 
\begin{equation*}
\begin{split}
    \hat{I}(p_i) & = I(p_i) + \gamma \mathbf{n} \frac{\partial \mathbb{F}_i}{\partial \theta^D} , \\
    \hat{I}(p_j) & = I(p_j) + \gamma \mathbf{n} \frac{\partial \mathbb{F}_j}{\partial \theta^D} .
\end{split}
\end{equation*}
We can now characterize the mutual information between $\hat{I}(p_i)$ and $\hat{I}(p_j)$ under the joint probability distribution $\mathbb{P}(\hat{I}(p_i), \hat{I}(p_j))$. As derived in JacobiNeRF, the mutual information can be computed as:
\begin{equation*}
    \begin{split}
        \mathbb{I}(\hat{I}(p_i),\hat{I}(p_j)) & =\mathbb{H}(\hat{I}(p_i))-\mathbb{H}(\hat{I}(p_i)|\hat{I}(p_j)) \\
        & = \log(\frac{1}{\sqrt{1-\cos^2\beta}}) + \text{const.}. \\
    \end{split}
\end{equation*}
Here, $\beta$ denotes the angle between 
$\partial \mathbb{F}_i / \partial \theta^D$ 
and 
$\partial \mathbb{F}_j / \partial \theta^D$. 
It is observed that: 
\begin{equation}
\label{equ:mutualinfomation}
\mathbb{I}(\hat{I}(p_i),\hat{I}(p_j)) \propto ||\cos\beta||.
\end{equation}
Therefore, to enhance the MI between two random pixels induced by NeRF perturbations, 
one can enforce the absolute value of the cosine similarity. 
That is, if two pixels $p_i$, $p_j$ come with high correlation (e.g., from the same object in the scene), their gradients regarding the perturbed parameters should be in the same or opposite direction.

\subsection{Mutual Information Shaping of Geometry in SDF-based NeRFs} 
\label{sec:sdfshaping}
While the previous approach focuses on enforcing mutual information for image synthesis, 
we explore a mutual information shaping of the geometries for quality surface reconstruction in SDF-based NeRFs. 

\begin{figure*}[tb]
\centering
\includegraphics[width=1.0\linewidth]{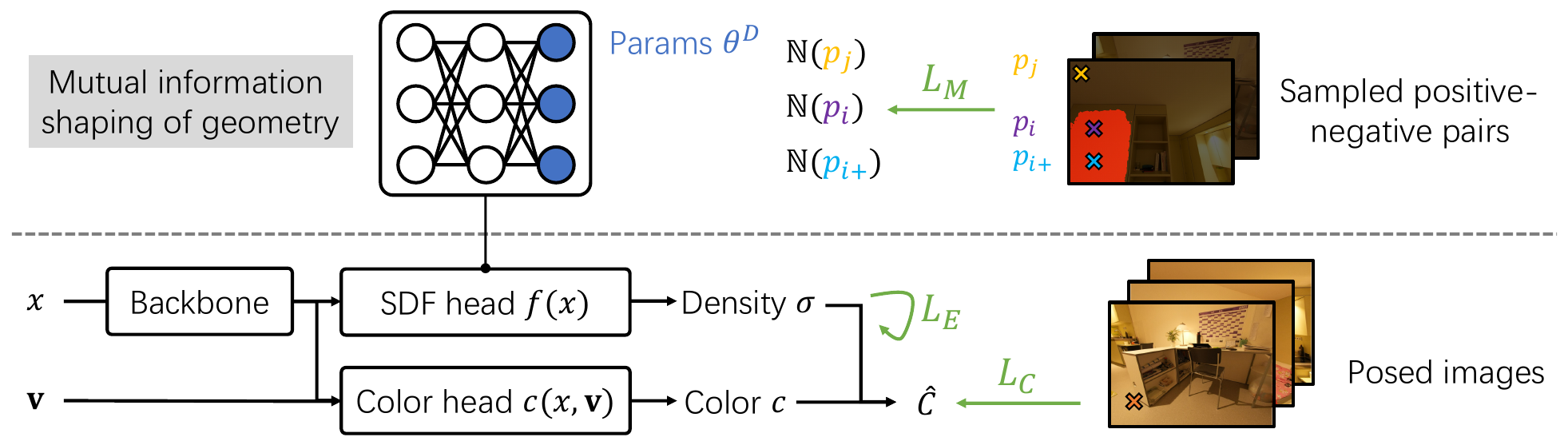}
\caption{
An {\bf overview} of the pipeline, 
where we apply mutual information shaping on the geometric branch to enforce consistencies that help enhance the surface reconstruction.
Specifically, the model consists of the NeRF backbone, an SDF head, and a color head. 
The predicted density and color are supervised by the classic eikonal loss $L_E$ and the photometric reconstruction loss $L_C$, respectively. 
We encode geometry-aware mutual information into a subset of the parameters $\theta^D$ of $f(x)$ to constrain the density field learning for better surface quality, which is achieved by the proposed mutual information loss $L_M$ computed on top of the estimated surface normal. 
}
\label{fig:pipeline}
\end{figure*}

Specifically, we represent a scene via a NeRF backbone with an SDF head and a color head, 
as shown in Fig.~\ref{fig:pipeline}. 
Note that the method to be proposed is capable of being applied to any network similar to NeRF and SDF representations. 
Given that the two heads partially share parameters, 
we denote all the parameters as $\theta$, and use 
$f(x;\theta)$ and $c(x, \mathbf{v}; \theta)$ to represent the SDF and color branches, respectively. 
To achieve mutual information shaping of the geometry, 
we focus on manipulating the parameters of $f(x;\theta)$, while keeping $c(x,\mathbf{v};\theta)$ unchanged. 
Below, we first explain why and how we compute mutual information with surface normals derived from SDF, i.e., $f(x;\theta)$, as a geometry-aware constraint. 
Next, we present a simple yet effective strategy that extracts regions with high correlation, which need to be encoded into the weights using the mutual information shaping technique.
Finally, we elaborate on our loss function and training process to reconstruct the scene from posed images. 

\paragraph{Mutual information among surface normals.}
We explicitly encode mutual information into the function $f(x;\theta)$. 
However, instead of computing mutual information between the density values, 
we proceed by manipulating the surface normal vectors that can be derived from $f$. 
Specifically, for a 3D point $x$ on a surface of the scene, its normal vector $N(x;\theta)$ can be calculated as the partial derivative of the 3D coordinate:
\begin{equation}
    N(x;\theta)=\frac{\partial f(x;\theta)}{\partial x}.
\end{equation}
There are several reasons to manipulate surface normal vectors but not SDF values: 
1) Shaping the SDF values with respect to a subset $\theta^D$ of the parameters may not effectively impose the constraints on all parameters. 
However, the calculation of the surface normal includes the entire set of parameters in $f$, thus a more global regularity.
2) Moreover, shaping $f$ with the mutual information between its values may induce side effects as density could be distorted to satisfy the constraints. However, shaping the normal $\partial f / \partial x$ leave more space for the SDF values to cope with the scene reconstruction task, thus minimizing potential negative effects.

According to Eq.~\ref{equ:mutualinfomation}, 
we need to calculate the partial gradients $\partial N/ \partial \theta^D$ to compute the mutual information. 
When it comes to a vector rather than a scalar, the gradients consist of three components corresponding to the \textbf{x}, \textbf{y}, and \textbf{z} axes:  
\begin{equation}
    \frac{\partial N(x; \theta)}{\partial \theta^D} = \frac{\partial N_\textbf{x}(x; \theta)}{\partial \theta^D}, \frac{\partial N_\textbf{y}(x; \theta)}{\partial \theta^D}, \frac{\partial N_\textbf{z}(x; \theta)}{\partial \theta^D}.
\end{equation}
For ease of computation, we concatenate the gradients into a vector, denoted as $\partial N$ for simplicity:
\begin{equation}
    \partial N = \text{Concat}(\frac{\partial N_\textbf{x}} {\partial \theta^D}, \frac{\partial N_\textbf{y}} {\partial \theta^D}, \frac{\partial N_\textbf{z}} {\partial \theta^D}).
\end{equation}
Since a 2D pixel is accumulated from 3D, 
we aggregate the normal-based information along a viewing ray as in Eq.~\ref{equ:volumerendering} as: 
\begin{equation}
    \mathbb{N}(p_i) = \sum_{x\in \text{ray}}  \omega(t_i; \theta)\partial N .
\end{equation}
Finally, we obtain that the mutual information between two normals is: 
\begin{equation}
\label{eq:normal_mi}
    \mathbb{I}(\hat{N}(p_i), \hat{N}(p_j)) \propto ||\frac{\mathbb{N}(p_i) \cdot \mathbb{N}(p_j)}{|\mathbb{N}(p_i)||\mathbb{N}(p_j)|}|| .
\end{equation}
Please refer to our supplementary material for a detailed derivation. 
From Eq.~\ref{eq:normal_mi}, we observe that the mutual information in surface normal vectors correlates positively with the cosine similarity of their gradients to the network parameters. This conclusion is similar to that of Eq.~\ref{equ:mutualinfomation}. 
As a result, we can now identify the pixels that are of high correlations, 
and increase the aggregated mutual information during the NeRF training process.

\paragraph{Extracting correlated scene regions from image pixels.}
To extract scene regions that need to be shaped to have high mutual information, 
we present a simple yet effective approach with off-the-shelf visual features. 
Given the input images, we extract pixel-wise semantic and geometric features to identify correlated scene regions. 
Following previous research~\cite{xu2023jacobinerf}, we apply DINO~\cite{caron2021emerging} as the semantic feature extractor to obtain a high-level understanding of the scene regions. 
Meanwhile, as we shape mutual information on top of the surface normal, we employ a monocular normal estimator~\cite{do2020surface} as the geometric feature extractor for a detailed low-level perception of the 3D points in the scene.

\begin{figure}[tb]
\centering
\includegraphics[width=\linewidth]{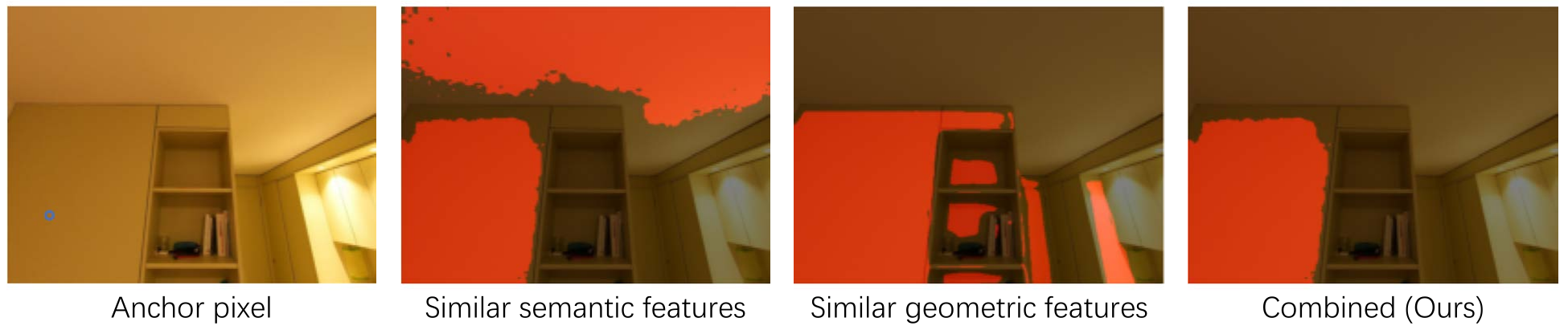}
\caption{
An example of positive samples from different features. 
Given an anchor pixel (marked by a blue circle in the first image) on the wall-like cabinet, semantic features like DINO often correlate both the cabinet and the ceiling (as shown by the red area in the second image). Meanwhile, geometric features such as (noisy) monocular normals can not distinguish between parallel planes that are not connected (the third image). 
By combining semantic and geometric features, 
we can obtain positive samples of the anchor pixel with better geometric consistency, thus, high mutual correlation.
}
\label{fig:positive}
\end{figure}

DINO features can group different parts of an object, like the seat and back of a chair, which share similar semantics. 
However, the surface normal vectors from these parts have limited correlation, thus should not be constrained. 
Conversely, using only geometric features can wrongly suggest high correlation between unrelated areas, like the seat and the floor. 
As these areas aren't closely connected in 3D space, there is no reason to have better surface reconstruction when enhancing their mutual information. 
Similar phenomena on wall-like structures can be observed in Fig.~\ref{fig:positive}, 
where we show that by combining semantic and geometric features together we can better identify scene regions that are spatially connected and have similar normal directions.

More explicitly, 
we combine semantic and geometric features with an intersection operator. 
It is simple yet effective. 
The process of identifying correlated scene regions (in normal) is implemented as positive pixel pair selection from input images. 
Given a randomly selected pixel from an image, we first extract similar pixels from neighboring images using semantic and geometric features, respectively. 
This extraction is implemented by applying two thresholds, $\beta_{S}$ and $\beta_{G}$, to filter the cosine similarity for semantic and geometric features. 
We then feed the intersection of the top similar pixels from each feature type into NeRF's training process as positive samples, allowing us to shape them with high mutual information. 
In order to leverage the contrastive learning framework to encourage mutual information, we randomly select pixels not involved in the aforementioned process to serve as negative samples (independent pixels that are not correlated to the anchor pixel).

\paragraph{Loss functions for training.} 
The aforementioned contrastive learning process is performed with a tailored InfoNCE loss~\cite{oord2018representation}:
\begin{equation}
\label{equ:infonce}
    L_M = -\log\frac{\sum \exp(||\cos(\partial\mathbb{N}_i, \partial\mathbb{N}_{i+})||)}{\sum_j \exp(||\cos(\partial\mathbb{N}_i, \partial\mathbb{N}_j)||)},
\end{equation}
where $i$, $i+$, and $j$ denote the anchor, positive, and negative pixels, respectively. 
This loss function effectively enhances mutual information (according to Eq.~\ref{eq:normal_mi}) among correlated pixels, while maintaining low mutual information among independent pixels.
Finally, to train the NeRF backbone and retain the characteristics of the SDF, we adopt the basic photometric loss (Eq.~\ref{eq:photometric_loss}) and the eikonal loss (Eq.~\ref{eq:eikonal_loss}). 
The latter two are derived from single images, while the former can be calculated from either single images or a set of neighboring images. 
The full loss function is defined as the weighted sum of three loss terms: 
\begin{equation}
\label{equ:totalloss}
    L = L_C + \lambda_{E} L_E + \lambda_{M} L_M .
\end{equation}
With the above, all network parameters $\theta$ can be trained end-to-end.

\section{Experiments}

In this section, 
we evaluate the effectiveness of our method through experiments conducted on multiple datasets with various baselines. 
First, we introduce the two datasets used to conduct the experiments in Sec.~\ref{sec:dataset}. 
Next, we detail the implementation of the proposed method in Sec.~\ref{sec:implementation}, 
where we adapt a spectrum of state-of-the-art models for validating the effectiveness of the geometric shaping.
Then, we report the comparisons between the baselines and our models in Sec.~\ref{sec:comparisons}. 
Lastly, we conduct a series of ablation studies in Sec.~\ref{sec:analyses}, to verify the necessity of each component in our method.

\subsection{Datasets}
\label{sec:dataset} 
We evaluate the proposed method using two public datasets from the literature: ScanNet++~\cite{yeshwanthliu2023scannetpp} and the Replica~\cite{straub2019replica} dataset. 
For the ScanNet++ dataset, we choose 8 representative scenes, while for the Replica dataset, we select 4 representative scenes. The scenes from both datasets represent a variety of indoor settings, including offices, bedrooms, and so on. 
As our focus is on scene reconstruction from sparse views, we uniformly sample 50-85 images from the official (and Replica from Semantic-NeRF~\cite{zhi2021place}) scanning sequences for each scene to use as input in our evaluation. 
For detailed statistics of each scene, please refer to our supplementary material.

\paragraph{Evaluation metrics.}
Following previous works~\cite{wang2021neus,yu2022monosdf,li2023neuralangelo}, 
we evaluate the quality of surface reconstruction using two metrics: Chamfer distance and F-score. 
Chamfer distance is a measure that evaluates the difference between reconstructed and ground-truth points sets. 
F-score is a statistical metric that integrates precision and recall, providing a more comprehensive understanding of the reconstruction quality. 

\subsection{Implementation Details}
\label{sec:implementation}

As the proposed geometric shaping can be easily integrated into any network similar to NeRF with SDF representations, 
we select several state-of-the-art models as our baselines and apply the proposed to them for demonstrating the effectiveness of our approach. 
We first choose NeuS~\cite{wang2021neus} and VolSDF~\cite{yariv2021volume}, the two representative pioneers for NeRFs with SDFs. 
We also apply our method to the following works: 
GeoNeuS~\cite{fu2022geo},
I$^2$-SDF~\cite{zhu2023i2sdf}, 
NeuRIS~\cite{wang2022neuris}, 
MonoSDF~\cite{yu2022monosdf}, 
and Neuralangelo~\cite{li2023neuralangelo}.
In addition to the basic loss functions for NeRF and SDF, 
GeoNeuS incorporates patch similarity from neighboring views and surface points derived from structure-from-motion.
I$^2$-SDF integrates the rendering equation with VolSDF. 
We disregard its normal and depth loss terms. 
NeuRIS utilizes monocular normal estimation as a supervision on the geometry, and discards the low-quality normal information that lacks sufficient confidence through multi-view patching.
MonoSDF employs both depth and normal estimation, for which we set the decay of the normal and depth at 30k iterations to prevent the method from degeneration. 
Neuralangelo utilizes a multi-resolution hash grid and numerical gradient. 
We set the hash encoding dictionary size to 20, and the encoding dimension to 4.
For all the base models mentioned above, 
we enhance them by incorporating our proposed mutual information loss during training. 
The resulting methods are respectively labeled as NeuS$^+$, VolSDF$^+$, GeoNeuS$^+$, and so on. 

\subsection{Comparisons}
\label{sec:comparisons}

\begin{table*}[th!]

\caption{
Quantitative results on the ScanNet++ dataset. 
Each cell includes the original baseline number in black, followed by the improvement using our mutual information. 
Positive and negative improvements are marked in \textcolor{mygreen}{green} and \textcolor{orange}{orange}, respectively.
The last column shows the average improvement over all scenes, validating the effectiveness of the proposed shaping for better geometric reconstruction (despite a small potion of noise on a few scenes).
}

\centering
\resizebox{1.0\textwidth}{!}{ 
\begin{tabular}{cl|cccccccc|c}

\toprule
 & & 0a7c & 0a18 & 6ee2 & 7b64 & 56a0 & 9460 & a08d & e0ab & Mean \\ 

\midrule

\multirow{7}{*}{\rotatebox{90}{Chamfer (m)$\downarrow$}}  
& NeuS~\cite{wang2021neus}$^+$ & 0.08{\textcolor{mygreen}{-0.04}} & 0.05{\textcolor{mygreen}{-0.03}} & 0.03{\textcolor{orange}{+0.01}} & 0.06{\textcolor{mygreen}{-0.04}} & 0.01{\textcolor{mygreen}{-0.00}} & 0.24{\textcolor{mygreen}{-0.14}} & 0.04{\textcolor{mygreen}{-0.03}} & 0.02{\textcolor{mygreen}{-0.00}} & 0.07{\textcolor{mygreen}{-0.03}} \\ 

& VolSDF~\cite{yariv2021volume}$^+$ & 0.90\textcolor{mygreen}{-0.83} & 0.31\textcolor{mygreen}{-0.29} & 0.04\textcolor{orange}{+0.02} & 0.10\textcolor{mygreen}{-0.03} & 0.03\textcolor{mygreen}{-0.01} & 0.14\textcolor{orange}{+0.03} & 0.39\textcolor{mygreen}{-0.38} & 0.04\textcolor{mygreen}{-0.00} & 0.24\textcolor{mygreen}{-0.19}\\ 

& GeoNeuS~\cite{fu2022geo}$^+$ & 0.03\textcolor{mygreen}{-0.02} & 1.03\textcolor{mygreen}{-1.02} & 0.28\textcolor{mygreen}{-0.27} & 0.05\textcolor{mygreen}{-0.03} & 0.02\textcolor{mygreen}{-0.01} & 0.22\textcolor{mygreen}{-0.20} & 0.01\textcolor{mygreen}{-0.01} & 0.26\textcolor{mygreen}{-0.25} & 0.24\textcolor{mygreen}{-0.23}\\ 

& I$^2$-SDF~\cite{zhu2023i2sdf}$^+$ & 0.27\textcolor{mygreen}{-0.24} & 0.22\textcolor{mygreen}{-0.03} & 0.04\textcolor{mygreen}{-0.01} & 0.47\textcolor{mygreen}{-0.45} & 0.03\textcolor{mygreen}{-0.02} & 0.13\textcolor{orange}{+0.05} & 0.03\textcolor{orange}{+0.02} & 0.30\textcolor{mygreen}{-0.07} & 0.19\textcolor{mygreen}{-0.09} \\ 

& NeuRIS~\cite{wang2022neuris}$^+$ & 0.01\textcolor{mygreen}{-0.00} & 0.01\textcolor{mygreen}{-0.00} & 0.03\textcolor{orange}{+0.02} & 0.07\textcolor{mygreen}{-0.04}& 0.04\textcolor{mygreen}{-0.03}  & 0.41\textcolor{mygreen}{-0.21} & 0.03\textcolor{mygreen}{-0.01} & 0.03\textcolor{mygreen}{-0.00} & 0.08\textcolor{mygreen}{-0.03}\\ 

& MonoSDF~\cite{yu2022monosdf}$^+$ & 0.01\textcolor{mygreen}{-0.00} & 0.01\textcolor{orange}{+0.02} & 0.02\textcolor{mygreen}{-0.01} & 0.02\textcolor{mygreen}{-0.00} & 0.01\textcolor{mygreen}{-0.00} & 0.05\textcolor{mygreen}{-0.02} & 0.01\textcolor{mygreen}{-0.00} & 0.01\textcolor{mygreen}{-0.00}  & 0.02\textcolor{mygreen}{-0.00}\\ 

& Neuralangelo~\cite{li2023neuralangelo}$^+$ & 0.17\textcolor{mygreen}{-0.01} & 0.84\textcolor{mygreen}{-0.66} & 0.52\textcolor{mygreen}{-0.33} & 0.05\textcolor{mygreen}{-0.01} & 0.04\textcolor{orange}{+0.02} & 4.76\textcolor{mygreen}{-2.85} & 0.33\textcolor{orange}{+0.05} & 0.12\textcolor{orange}{+0.03}  & 0.85\textcolor{mygreen}{-0.48} \\ 

\midrule

\multirow{7}{*}{\rotatebox{90}{F-score $\uparrow$}} 
& NeuS~\cite{wang2021neus}$^+$ & 0.49\textcolor{mygreen}{+0.06} & 0.48\textcolor{mygreen}{+0.10} & 0.67\textcolor{mygreen}{+0.00} & 0.58\textcolor{mygreen}{+0.16} & 0.81\textcolor{orange}{-0.01} & 0.58\textcolor{orange}{-0.02} & 0.67\textcolor{mygreen}{+0.17} & 0.75\textcolor{mygreen}{+0.09} & 0.63\textcolor{mygreen}{+0.07} \\ 

& VolSDF~\cite{yariv2021volume}$^+$ & 0.45\textcolor{mygreen}{+0.35} & 0.47\textcolor{mygreen}{+0.27} & 0.67\textcolor{mygreen}{+0.00} & 0.55\textcolor{mygreen}{+0.05} & 0.71\textcolor{mygreen}{+0.06} & 0.52\textcolor{mygreen}{+0.18} & 0.48\textcolor{mygreen}{+0.35} & 0.70\textcolor{mygreen}{+0.02} & 0.57\textcolor{mygreen}{+0.16} \\ 

& GeoNeuS~\cite{fu2022geo}$^+$ & 0.69\textcolor{mygreen}{+0.20} & 0.47\textcolor{mygreen}{+0.44} & 0.52\textcolor{mygreen}{+0.34} & 0.70\textcolor{mygreen}{+0.06} & 0.87\textcolor{mygreen}{+0.03} & 0.68\textcolor{mygreen}{+0.23} & 0.75\textcolor{mygreen}{+0.14} & 0.37\textcolor{mygreen}{+0.51} & 0.63\textcolor{mygreen}{+0.24} \\ 

& I$^2$-SDF~\cite{zhu2023i2sdf}$^+$ & 0.33\textcolor{mygreen}{+0.52} & 0.28\textcolor{mygreen}{+0.00} & 0.58\textcolor{mygreen}{+0.15} & 0.41\textcolor{mygreen}{+0.43} & 0.75\textcolor{mygreen}{+0.10} & 0.65\textcolor{mygreen}{+0.07} & 0.65\textcolor{mygreen}{+0.07} & 0.40\textcolor{mygreen}{+0.25} & 0.51\textcolor{mygreen}{+0.20} \\ 

& NeuRIS~\cite{wang2022neuris}$^+$ & 0.72\textcolor{mygreen}{+0.03} & 0.68\textcolor{mygreen}{+0.01} & 0.65\textcolor{mygreen}{+0.01} & 0.69\textcolor{mygreen}{+0.06} & 0.85\textcolor{mygreen}{+0.01} & 0.24\textcolor{mygreen}{+0.17} & 0.64\textcolor{mygreen}{+0.13} & 0.69\textcolor{mygreen}{+0.09} & 0.65\textcolor{mygreen}{+0.06} \\ 

& MonoSDF~\cite{yu2022monosdf}$^+$ & 0.92\textcolor{orange}{-0.01} & 0.88\textcolor{orange}{-0.01} & 0.81\textcolor{mygreen}{+0.04} & 0.79\textcolor{mygreen}{+0.06} & 0.89\textcolor{mygreen}{+0.00} & 0.88\textcolor{mygreen}{+0.05} & 0.91\textcolor{mygreen}{+0.00} & 0.89\textcolor{orange}{-0.01} & 0.87\textcolor{mygreen}{+0.01} \\ 

& Neuralangelo~\cite{li2023neuralangelo}$^+$ & 0.42\textcolor{mygreen}{+0.05} & 0.14\textcolor{mygreen}{+0.30} & 0.35\textcolor{mygreen}{+0.12} & 0.63\textcolor{mygreen}{+0.05} & 0.71\textcolor{orange}{-0.01} & 0.17\textcolor{mygreen}{+0.22} & 0.29\textcolor{mygreen}{+0.06} & 0.51\textcolor{orange}{-0.01} & 0.40\textcolor{mygreen}{+0.10} \\ 

\bottomrule

\end{tabular}

}

\label{tab:scannetpp}
\end{table*}
\begin{table*}[!t]
\caption{
Quantitative results on the Replica dataset.
}

\centering
\resizebox{0.75\textwidth}{!}{ 
\begin{tabular}{cl|cccc|c}

\toprule
 & & office0 & office1 & room0 & room1 & Mean \\ 

\midrule

\multirow{7}{*}{\rotatebox{90}{Chamfer (m)$\downarrow$}}  
& NeuS~\cite{wang2021neus}$^+$ & 0.005\textcolor{orange}{+0.001} & 0.023\textcolor{mygreen}{-0.012} & 0.068\textcolor{mygreen}{-0.038} & 0.130\textcolor{mygreen}{-0.087} & 0.057\textcolor{mygreen}{-0.035} \\ 

& VolSDF~\cite{yariv2021volume}$^+$ &0.004\textcolor{orange}{+0.006} & 0.016\textcolor{mygreen}{-0.004} & 0.080\textcolor{mygreen}{-0.071} & 0.034\textcolor{mygreen}{-0.025} & 0.033\textcolor{mygreen}{-0.023}\\ 

& GeoNeuS~\cite{fu2022geo}$^+$ & 0.023\textcolor{mygreen}{-0.021} & 0.014 \textcolor{mygreen}{-0.009}  & 0.040 \textcolor{mygreen}{-0.035}  & 0.002 \textcolor{mygreen}{-0.000}  & 0.020\textcolor{mygreen}{-0.017}  \\ 

& I$^2$-SDF~\cite{zhu2023i2sdf}$^+$ & 0.032\textcolor{mygreen}{-0.009} & 0.056\textcolor{mygreen}{-0.033} & - & - & 0.044\textcolor{mygreen}{-0.021} \\ 

& NeuRIS~\cite{wang2022neuris}$^+$ &  0.014\textcolor{mygreen}{-0.006} & 3.975\textcolor{mygreen}{-3.963} & 0.025\textcolor{mygreen}{-0.017} & 0.015\textcolor{orange}{+0.002} & 1.007\textcolor{mygreen}{-0.996} \\ 

& MonoSDF~\cite{yu2022monosdf}$^+$ & 0.003\textcolor{mygreen}{-0.000} & 0.005 \textcolor{mygreen}{-0.001}  & 0.004 \textcolor{mygreen}{-0.000}  & 0.004 \textcolor{mygreen}{-0.001}  & 0.004\textcolor{mygreen}{-0.001}\\ 

& Neuralangelo~\cite{li2023neuralangelo}$^+$ & 0.024\textcolor{mygreen}{-0.018} & 0.030\textcolor{mygreen}{-0.019} & 0.146\textcolor{mygreen}{-0.012} & 0.082\textcolor{mygreen}{-0.047} & 0.070\textcolor{mygreen}{-0.024} \\ 

\midrule

\multirow{7}{*}{\rotatebox{90}{F-score $\uparrow$}} 
& NeuS~\cite{wang2021neus}$^+$ & 0.92\textcolor{mygreen}{+0.00} & 0.80\textcolor{mygreen}{+0.01} & 0.78\textcolor{orange}{-0.03} & 0.72\textcolor{mygreen}{+0.08} & 0.80\textcolor{mygreen}{+0.02}  \\ 

& VolSDF~\cite{yariv2021volume}$^+$ & 0.94\textcolor{orange}{-0.04} & 0.78\textcolor{mygreen}{+0.01} & 0.78\textcolor{mygreen}{+0.16} & 0.82\textcolor{mygreen}{+0.07} & 0.83\textcolor{mygreen}{+0.05}  \\ 

& GeoNeuS~\cite{fu2022geo}$^+$ & 0.89\textcolor{mygreen}{+0.08} & 0.91 \textcolor{mygreen}{+0.04}  & 0.90 \textcolor{mygreen}{+0.09}  & 0.98 \textcolor{mygreen}{+0.01}  & 0.920\textcolor{mygreen}{+0.05} \\ 

& I$^2$-SDF~\cite{zhu2023i2sdf}$^+$ & 0.72\textcolor{mygreen}{+0.09} & 0.56\textcolor{mygreen}{+0.20} &  - & - &  0.64\textcolor{mygreen}{+0.14}                        \\ 

& NeuRIS~\cite{wang2022neuris}$^+$ & 0.79\textcolor{mygreen}{+0.06} & 0.01\textcolor{mygreen}{+0.68} & 0.89\textcolor{mygreen}{+0.05} & 0.77\textcolor{mygreen}{+0.03} & 0.61\textcolor{mygreen}{+0.20} \\ 

& MonoSDF~\cite{yu2022monosdf}$^+$ & 0.96\textcolor{mygreen}{+0.00} & 0.90 \textcolor{mygreen}{+0.02}  & 0.98 \textcolor{mygreen}{+0.00}  & 0.95 \textcolor{mygreen}{+0.02}  & 0.95\textcolor{mygreen}{+0.01} \\ 

& Neuralangelo~\cite{li2023neuralangelo}$^+$ & 0.89\textcolor{mygreen}{+0.07} & 0.49\textcolor{mygreen}{+0.36} & 0.65\textcolor{mygreen}{+0.15} & 0.71\textcolor{mygreen}{+0.18} & 0.69\textcolor{mygreen}{+0.19} \\ 

\bottomrule

\end{tabular} 
}

\label{tab:replica}
\end{table*}

\begin{figure}[!tbh]
\centering
\includegraphics[width=0.99\linewidth]{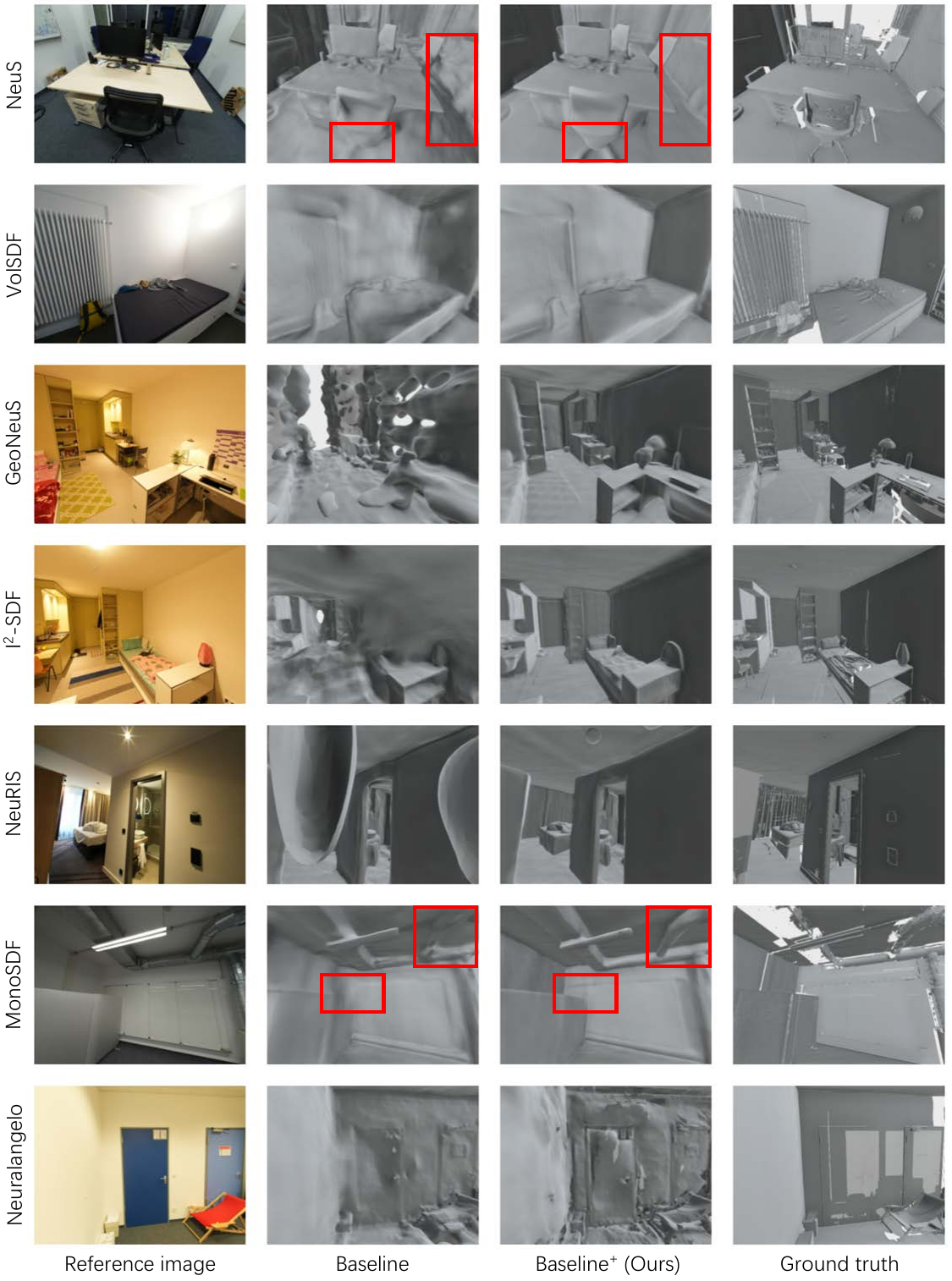}
\caption{
Visual results on the ScanNet++ dataset. 
Each row shows a comparison with a different baseline, from top to bottom,
NeuS~\cite{wang2021neus}$^+$, VolSDF~\cite{yariv2021volume}$^+$, GeoNeuS~\cite{fu2022geo}$^+$, I$^2$-SDF~\cite{zhu2023i2sdf}$^+$, NeuRIS~\cite{wang2022neuris}$^+$, MonoSDF~\cite{yu2022monosdf}$^+$, and Neuralangelo~\cite{li2023neuralangelo}$^+$. 
Red boxes are overlaid to help the contrast.
}
\label{fig:scannetpp}
\end{figure}

\paragraph{Results on the ScanNet++ dataset.}
The quantitative comparisons are reported in Tab.~\ref{tab:scannetpp}. 
We also visualize the surface reconstruction results in Fig.~\ref{fig:scannetpp} for qualitative comparisons.  
We can observe that by applying the proposed mutual information shaping on the geometry, the surface reconstruction quality can be improved across all baselines, as measured by mean Chamfer distance, F-score, and the visuals.

NeuS and VolSDF tend to produce rough surfaces and excessively smoothed boundaries. 
However, with our geometry-aware mutual information, the surfaces become smoother, while the boundaries are more crispy. 
Compared to GeoNeuS, a method that uses patch-similarity among neighboring views and requires sufficient textures, our method provides significant improvements. 
Applying the proposed technique to I$^2$-SDF can also enhance its reconstruction quality.

As for more recent works, 
we find that NeuRIS is somewhat sensitive to the quality of normal estimation.
It can either be affected by noise in the normal or completely disregard the normal information due to low confidence. 
Particularly, when images are sparse, the regions covered by fewer viewpoints tend to lose supervision.
Our mutual information scheme does not depend on the absolute values of the surface normal, but on the similarity among these values. 
This makes it more robust to noise and less likely to discard low-quality but useful normal information. 
Consequently, our method enhances the reconstruction quality of NeuRIS.
MonoSDF achieves the best overall performance among the baselines because it uses both depth and normal estimations as geometry supervision. 
However, our method still offers improvements. Notably, it produces smoother planes and clearer boundaries.
This indicates that not only the supervision of geometric cues is important, but also the correlation between the geometric entities matters.

Last but not least, 
the experiments on Neuralangelo demonstrate the potential of our method in working with hash grid encoding and numerical gradient. 
This highlights our method's versatility and ease of use, which can be broadly applied in various reconstruction architectures. 

\paragraph{Results on the Replica dataset.}
In Tab.~\ref{tab:replica} we report the numerical results on the Replica dataset, with respect to NeuS~\cite{wang2021neus}, 
VolSDF~\cite{yariv2021volume}, 
GeoNeuS~\cite{fu2022geo},
I$^2$-SDF~\cite{zhu2023i2sdf}, 
NeuRIS~\cite{wang2022neuris}, 
MonoSDF~\cite{yu2022monosdf}, 
and Neuralangelo~\cite{li2023neuralangelo}.
As observed, 
the proposed geometric shaping technique still provides positive improvements when incorporated into these baselines, aligning with the conclusion drawn from the ScanNet++ data.
Qualitative examples can also be found in Fig.~\ref{fig:teaser}.

\subsection{Analyses}
\label{sec:analyses}
\begin{wraptable}{r}{0.6\textwidth}

\caption{
Ablation studies on three scenes from ScanNet++.
The best and second-best results are marked in \textbf{bold} and \underline{underlined}.
}

\centering
\resizebox{0.6\textwidth}{!}{ 
\begin{tabular}{l|cccc|cccc}

\toprule

& \multicolumn{4}{|c|}{Chamfer (m)$\downarrow$} 
& \multicolumn{4}{|c}{F-score $\uparrow$} \\
 & 6ee2 & 7b64 & 9460 & Mean & 6ee2 & 7b64 & 9460 & Mean \\ 

\midrule

NeuRIS~\cite{wang2021neus} 
& 0.029 & 0.070 & 0.405 & 0.168 & 0.65 & 0.69 & 0.24 & 0.53 \\ 

NeuRIS$^+$ (color) & \textbf{0.023} & 0.038 & \underline{0.315} & \underline{0.125} & \textbf{0.72} & 0.68 & \underline{0.34} & \underline{0.58} \\ 
NeuRIS$^+$ (SDF) & 0.127 & 0.053 &  0.360 & 0.180 & 0.47 & 0.61 & 0.30 & 0.46 \\ 
NeuRIS$^+$ w/o normal & \underline{0.036} & \textbf{0.025} & 0.338 & 0.133 & 0.66 & \underline{0.75} & 0.32 & 0.58 \\ 
NeuRIS$^+$ w/o DINO & 0.061 & 0.033 & 0.381 & 0.158 & 0.63 & 0.72 & 0.32 & 0.56 \\ 
NeuRIS$^+$ (Full) & 0.044 & \underline{0.033} & \textbf{0.198} & \textbf{0.092} & \underline{0.66} & \textbf{0.76} & \textbf{0.41} & \textbf{0.61}  \\ 

\midrule

MonoSDF~\cite{yu2022monosdf} & \underline{0.020} & \underline{0.016} & 0.046 & 0.028 & \underline{0.81} & 0.79 & 0.88 & 0.83 \\ 
MonoSDF$^+$ (color) &  0.024 & 0.019 & 0.033 & \underline{0.025} & 0.79 & 0.78 & 0.87 & 0.81  \\ 
MonoSDF$^+$ (SDF) & 0.588 & 0.029 & 3.443 & 1.353 & 0.38 & 0.65 & 0.21 & 0.42 \\ 
MonoSDF$^+$ w/o normal & 0.128 & 0.023 & 0.066 & 0.072 & 0.68 & 0.73 & 0.80 & 0.74\\ 
MonoSDF$^+$ w/o DINO & 0.064 & \textbf{0.016} & \textbf{0.022} & 0.034 & 0.78 & \underline{0.83} & \underline{0.93} & \underline{0.85} \\ 
MonoSDF$^+$ (Full) & \textbf{0.014} & 0.020 & \underline{0.023} & \textbf{0.019} & \textbf{0.85} & \textbf{0.85} & \textbf{0.93}  & \textbf{0.87} \\ 

\bottomrule

\end{tabular} 
}

\label{tab:ablation}
\end{wraptable}

\paragraph{Ablation studies.}
\begin{figure}[!t]
\centering
\includegraphics[width=\linewidth]{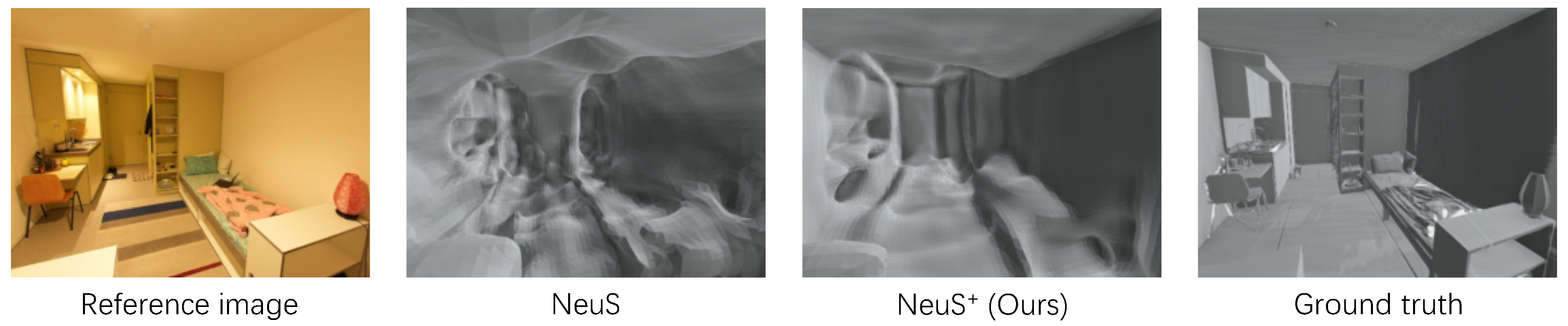}
\caption{
Reconstruction results after 10K iterations in the training process with (NeuS$^+$) and without (NeuS) the proposed geometric shaping technique. 
}
\label{fig:neus_10k}
\end{figure}

We select three representative scenes from the ScanNet++ dataset and conduct ablation studies to evaluate the effectiveness of our method designs. 
We conduct ablation studies using two top-performing baselines: NeuRIS and MonoSDF.

The results are reported in Tab.~\ref{tab:ablation}. 
Instead of using the surface normal formula to inject mutual information, we first perform a basic mutual information shaping separately in the color and SDF values. These two variations are labeled as base$^+$ (color) and base$^+$ (SDF), respectively.
Interestingly, we find that, even though color is not directly related to geometry, by shaping it, we can still obtain improvements on the geometry in some cases.
Moreover, shaping directly on the SDF values severely degrade the reconstruction quality, which evidences that the shaping on SDF values could incur a competition with the goal to faithfully learn the geometry within the neural radiance field.
We further examine the effectiveness of combining semantic and geometric features to identify geometrically highly correlated regions. 
As demonstrated in Fig.~\ref{fig:positive}, using only one type of the features can lead to incorrect correlations. 
Tab.~\ref{tab:ablation} also shows that eliminating any one of the features can decrease the performance, confirming the effectiveness of our positive-negative sample selection mechanism levering the multimodal features.

\begin{wraptable}{r}{0.50\textwidth}

\caption{
Time and memory consumption. 
}

\centering
\resizebox{0.50\textwidth}{!}{ 
\begin{tabular}{l|cc}

\toprule
& Training Time (h) & VRAM (GB) \\ 

\midrule

NeuS~\cite{wang2021neus}$^+$ & 3.00+0.37 & 7.89+1.68 \\ 

VolSDF~\cite{yariv2021volume}$^+$ & 3.18+1.96 & 16.36+4.68 \\ 

GeoNeuS~\cite{fu2022geo}$^+$ &3.26+0.16 & 5.72+3.56 \\ 

I$^2$-SDF~\cite{zhu2023i2sdf}$^+$ & 3.47+2.84  & 10.08+8.09 \\ 

NeuRIS~\cite{wang2022neuris}$^+$ & 3.70+0.20 & 11.73+4.12 \\ 

MonoSDF~\cite{yu2022monosdf}$^+$  &2.65+1.32  & 12.83+9.42 \\ 

Neuralangelo~\cite{li2023neuralangelo}$^+$ &3.12+2.06  & 16.44+1.00  \\ 

\bottomrule

\end{tabular} 
}

\label{tab:consume}
\end{wraptable}

\paragraph{Time and memory consumption.}
Our method provides benefits for 3D reconstruction with limited additional effort.
Specifically, it only increases the optimization time and memory usage during the training process, without affecting the model architecture and final storage.
The time and memory consumption are reported in Tab.~\ref{tab:consume}. 
It is observed that the overhead on the training time highly depends on the architecture used in each basline.
For example, the training time increase on NeuRIS is marginal while more significant on VolSDF.
These studies provide comprehensive information on the trade-offs between training time and performance gain.
However, we do find that baselines with the proposed shaping technique can achieve a better performance (e.g., early geometries from NeuS and NeuS$^+$ shown in Fig.~\ref{fig:neus_10k}) at the same training steps due to the effective regularization by mutual information.

\section{Conclusion}

We explore the usage of the second-order geometric correlations as a regularization and introduce a mutual information shaping technique of the normals for better surface reconstruction quality. 
With a simple yet effective correlated point selection mechanism that does not depend on accurate geometry information, we can perform a contrastive learning that maximizes the detected correlations.
In return, we can observe a boost in the geometric quality of the reconstructed surfaces in state-of-the-art neural scene representations.

Extensive evaluations on various baselines and datasets show the effectiveness of our method, which improves 3D reconstruction quality without resorting to a more sophisticated surface modeling.
The overhead caused by the shaping of the normals varies with different neural architectures of the scene representation, e.g., some are marginal, and some are significant, which we deem as a limitation of our current pipeline. We believe that an acceleration in computation can be achieved using a coarse-to-fine strategy, which we leave as a future work.

\clearpage
\noindent
\textbf{Acknowledgments.} 
We thank the anonymous reviewers for their valuable feedback. 
This work is supported by 
the Early Career Scheme of the Research Grants Council (grant \# 27207224),
the HKU-100 Award, 
and in part by NSF China (No. 62172363).
Siyan Dong would also like to thank the support from HKU Musketeers Foundation Institute of Data Science for the Postdoctoral Research Fellowship.

\section{Detailed Derivation of Mutual Information of Normals}

In this section, we derive the mutual information of normals and explain our approximation. 

The surface normals can be written as 
\begin{equation*}
    \begin{split}
        \mathbb{N}(p_i)=(\mathbb{N}_\textbf{x}^i,\mathbb{N}_\textbf{y}^i,\mathbb{N}_\textbf{z}^i) , \\
        \mathbb{N}(p_j)=(\mathbb{N}_\textbf{x}^j,\mathbb{N}_\textbf{y}^j,\mathbb{N}_\textbf{z}^j) .
    \end{split}
\end{equation*}
When perturbing the normals by a random noise $n \in \mathbb{R}^D$ sampled from $\mathbb{S}^{D-1}$, according to Talor expansion, we have 
\begin{equation*}
    \begin{split}
\hat{\mathbb{N}}(p_i) &= \mathbb{N}(p_i) + \gamma n \cdot \frac{\partial{\mathbb{F}(o_i, v_i; \theta^D+n)}}{\partial{\theta^D}} , \\
\hat{\mathbb{N}}(p_j) &= \mathbb{N}(p_j) + \gamma n \cdot \frac{\partial{\mathbb{F}(o_j, v_j; \theta^D+n)}}{\partial{\theta^D}} ,
    \end{split}
\end{equation*}
where $\mathbb{F}$ is the function to calculate the normal information along the ray (weighted sum of the parameter gradients).
Let $\partial{\mathbb{F}_i}/\partial{\theta^D} = (\gamma_{A_\textbf{x}}A_\textbf{x}, \gamma_{A_\textbf{y}}A_\textbf{y}, \gamma_{A_\textbf{z}}A_\textbf{z})$, and  $\partial{\mathbb{F}_j}/\partial{\theta^D} = (\gamma_{B_\textbf{x}}B_\textbf{x}, \gamma_{B_\textbf{y}}B_\textbf{y}, \gamma_{B_\textbf{z}}B_\textbf{z})$, 
where $\gamma$ denotes the length of the vector, and the normal's partial vectors are all unit vectors, i.e., $A_\textbf{x}, A_\textbf{y}, A_\textbf{z}, B_\textbf{x}, B_\textbf{y}, B_\textbf{z} \in \mathbb{R}^{D-1}$.
The computation of mutual information can be written as
\begin{equation*}
\begin{split}
\mathbb{I}(\hat{\mathbb{N}}(p_i), \hat{\mathbb{N}}(p_j)) = & \mathbb{H}(\hat{\mathbb{N}}(p_i)) - \mathbb{H}(\hat{\mathbb{N}}(p_i) \mid \hat{\mathbb{N}}(p_j)) \\ 
= & \mathbb{H}(\gamma n \cdot (\gamma_{A_\textbf{x}}A_\textbf{x}, \gamma_{A_\textbf{y}}A_\textbf{y}, \gamma_{A_\textbf{z}}A_\textbf{z})) \\ - & \mathbb{H}(\gamma n \cdot (\gamma_{A_\textbf{x}}A_\textbf{x}, \gamma_{A_\textbf{y}}A_\textbf{y}, \gamma_{A_\textbf{z}}A_\textbf{z}) \mid \gamma n \cdot (\gamma_{B_\textbf{x}}B_\textbf{x}, \gamma_{B_\textbf{y}}B_\textbf{y}, \gamma_{B_\textbf{z}}B_\textbf{z})) .
\end{split}
\end{equation*}

Starting from a simple situation, we compute the entropy of $\gamma_{A_\textbf{x}}\gamma n \cdot A_\textbf{x}$. 
The entropy of a distribution shifts by the logarithm of the scaling factor when it's scaled. Therefore, we have
\begin{equation*}
    \mathbb{H}(\gamma_{A_\textbf{x}}\gamma n \cdot A_\textbf{x}) = \log (\gamma_{A_\textbf{x}}\gamma) + \mathbb{H}(\mathbb{S}^{D-1}) ,
\end{equation*}
where $\mathbb{H}(\mathbb{S}^{D-1})$ is a constant.
Then, for the joint entropy of two partial vectors:
\begin{equation*}
    \mathbb{H}(\gamma_{A_\textbf{x}}\gamma n \cdot A_\textbf{x}, \gamma_{A_\textbf{y}}\gamma n \cdot A_\textbf{y}) = \mathbb{H}(\gamma_{A_\textbf{x}}\gamma n \cdot A_\textbf{x}) + \mathbb{H}(\gamma_{A_\textbf{y}}\gamma n \cdot A_\textbf{y} \mid \gamma_{A_\textbf{x}}\gamma n \cdot A_\textbf{x}), 
\end{equation*}
where
\begin{equation*}
    \mathbb{H}(\gamma_{A_\textbf{y}}\gamma n \cdot A_\textbf{y} \mid \gamma_{A_\textbf{x}}\gamma n \cdot A_\textbf{x}) = \int_s \mathbb{H}(\gamma_{A_\textbf{y}}\gamma n \cdot A_\textbf{y} \mid \gamma_{A_\textbf{x}}\gamma n \cdot A_\textbf{x} = s) p(s) ds .
\end{equation*}
When $\gamma_{A_\textbf{x}}\gamma n \cdot A_\textbf{x} = s$
\begin{align*}
\begin{split}
    \gamma_{A_\textbf{y}}\gamma n \cdot A_\textbf{y} = & \gamma_{A_\textbf{y}}\gamma( \langle A_\textbf{x}, A_\textbf{y} \rangle A_\textbf{x} + (A_\textbf{y} - \langle A_\textbf{x}, A_\textbf{y} \rangle A_\textbf{x})) \\
    & \cdot (\langle n, A_\textbf{x} \rangle A_\textbf{x} + (n - \langle n, A_\textbf{x} \rangle A_\textbf{x})) \\
    = & \gamma_{A_\textbf{y}}\gamma (\langle n, A_\textbf{x} \rangle \cdot \langle A_\textbf{x}, A_\textbf{y} \rangle  + (A_\textbf{y} -  \langle A_\textbf{x}, A_\textbf{y} \rangle  A_\textbf{x})\cdot (n -  \langle n, A_\textbf{x} \rangle  A_\textbf{x})) ,
\end{split}
\end{align*}
so that
\begin{align*}
    & \mathbb{H}(\gamma_{A_\textbf{y}}\gamma n \cdot A_\textbf{y} \mid \gamma_{A_\textbf{x}}\gamma n \cdot A_\textbf{x} = s) \\
    = & \mathbb{H}(\gamma_{A_\textbf{y}}\gamma * (\langle n, A_\textbf{x} \rangle \cdot \langle A_\textbf{x}, A_\textbf{y} \rangle \\
    & + (A_\textbf{y} - \langle A_\textbf{x}, A_\textbf{y} \rangle A_\textbf{x})\cdot (n - \langle n, A_\textbf{x} \rangle \cdot A_\textbf{x})) \mid \gamma_{A_\textbf{x}}\gamma n \cdot A_\textbf{x} = s) \\
    = & \log(\gamma_{A_\textbf{y}}\gamma \sin(A_\textbf{x}, A_\textbf{y}) \sin(n, A_\textbf{x})) + \mathbb{H}(\mathbb{S}^{D-2}) .
\end{align*}

We have
\begin{align*}
    & \mathbb{H}(\gamma_{A_\textbf{y}}\gamma n \cdot A_\textbf{y} \mid \gamma_{A_\textbf{x}}\gamma n \cdot A_\textbf{x}) \\
    = & \int_s \mathbb{H}(\gamma_{A_\textbf{y}}\gamma n \cdot A_\textbf{y} \mid \gamma_{A_\textbf{x}}\gamma n \cdot A_\textbf{x} = s) p(s) ds \\
    = & \int_s (\log(\gamma_{A_\textbf{y}}\gamma \sin(A_\textbf{x}, A_\textbf{y}) \sin(n, A_\textbf{x})) + \mathbb{H}(\mathbb{S}^{D-2})) p(s) ds \\
    = & \mathbb{H}(\mathbb{S}^{D-2}) + \log(\gamma_{A_\textbf{y}}\gamma \sin (A_\textbf{x},A_\textbf{y}) ) + \int_s \log (\sin(n, A_\textbf{x})) p(s) ds \\
    = & \log(\gamma_{A_\textbf{y}}\gamma  \sin (A_\textbf{x},A_\textbf{y}) ) + const.
\end{align*}
In more general terms, we denote $A_\textbf{y} - \langle A_\textbf{x}, A_\textbf{y} \rangle A_\textbf{x}$ as $P(A_\textbf{y}, A_\textbf{x})$, which represents subtracting the component in the $A_\textbf{x}$ direction from the $A_\textbf{y}$.
Therefore, the equation above can be written as 
\begin{equation*}
    \mathbb{H}(\gamma_{A_\textbf{y}}\gamma n \cdot A_\textbf{y} \mid \gamma_{A_\textbf{x}}\gamma n \cdot A_\textbf{x}) = \log(\gamma_{A_\textbf{y}}\gamma |P(A_\textbf{y}, A_\textbf{x})|) + const.
\end{equation*}
\begin{equation*}
\begin{split}
    & \mathbb{H}(\gamma_{A_\textbf{x}}\gamma n \cdot A_\textbf{x}, \gamma_{A_\textbf{y}}\gamma n \cdot A_\textbf{y}) \\ 
    = & \mathbb{H}(\gamma_{A_\textbf{x}}\gamma n \cdot A_\textbf{x}) + \mathbb{H}(\gamma_{A_\textbf{y}}\gamma n \cdot A_\textbf{y} \mid \gamma_{A_\textbf{x}}\gamma n \cdot A_\textbf{x}) \\
    = & \log (\gamma_{A_\textbf{x}}\gamma) + \log(\gamma_{A_\textbf{y}}\gamma |P(A_\textbf{y}, A_\textbf{x})|) + const.
\end{split}
\end{equation*}
Similarly, we can infer that
\begin{equation*}
\begin{split}
    & \mathbb{H}(\gamma_{A_\textbf{x}}\gamma n \cdot A_\textbf{x}, \gamma_{A_\textbf{y}}\gamma n \cdot A_\textbf{y}, \gamma_{A_\textbf{z}}\gamma n \cdot A_\textbf{z}) \\ 
    = & \log (\gamma_{A_\textbf{x}}\gamma) + \log(\gamma_{A_\textbf{y}}\gamma |P(A_\textbf{y}, A_\textbf{x})|) + \log(\gamma_{A_\textbf{z}}\gamma |P(A_\textbf{z}, (A_\textbf{x}, A_\textbf{y}))|) + const.
\end{split}
\end{equation*}
\begin{align*}
    & \mathbb{H}(\gamma_{A_\textbf{x}}\gamma n \cdot A_\textbf{x}, \gamma_{A_\textbf{y}}\gamma n \cdot A_\textbf{y}, \gamma_{A_\textbf{z}}\gamma n \cdot A_\textbf{z} \mid \gamma n \cdot \gamma_{B_\textbf{x}}\gamma B_\textbf{x}, \gamma_{B_\textbf{y}}\gamma B_\textbf{y}, \gamma_{B_\textbf{z}}\gamma B_\textbf{z}) \\ 
    = & \log (\gamma_{A_\textbf{x}}\gamma|P(A_\textbf{x},(B_\textbf{x}, B_\textbf{y}, B_\textbf{z}))|) + \log(\gamma_{A_\textbf{y}}\gamma |P(A_\textbf{y}, (A_\textbf{x},B_\textbf{x}, B_\textbf{y}, B_\textbf{z}))|) \\
    & + \log(\gamma_{A_\textbf{z}}\gamma |P(A_\textbf{z}, (A_\textbf{x}, A_\textbf{y}, B_\textbf{x}, B_\textbf{y}, B_\textbf{z}))|) + const.
\end{align*}
By combining them, we obtain
\begin{equation*}
\begin{split}
    &\mathbb{I}(\hat{\mathbb{N}}(p_i), \hat{\mathbb{N}}(p_j)) \\
    = & \mathbb{H}(\hat{\mathbb{N}}(p_i)) - \mathbb{H}(\hat{\mathbb{N}}(p_i) \mid \hat{\mathbb{N}}(p_j)) \\
    = & \mathbb{H}(\gamma n \cdot (\gamma_{A_\textbf{x}}A_\textbf{x}, \gamma_{A_\textbf{y}}A_\textbf{y}, \gamma_{A_\textbf{z}}A_\textbf{z})) - \\ & \mathbb{H}(\gamma n \cdot (\gamma_{A_\textbf{x}}A_\textbf{x}, \gamma_{A_\textbf{y}}A_\textbf{y}, \gamma_{A_\textbf{z}}A_\textbf{z}) \mid \gamma n \cdot (\gamma_{B_\textbf{x}}B_\textbf{x}, \gamma_{B_\textbf{y}}B_\textbf{y}, \gamma_{B_\textbf{z}}B_\textbf{z})) \\
    = & \log (\gamma_{A_\textbf{x}}\gamma) \\
    & + \log(\gamma_{A_\textbf{y}}\gamma |P(A_\textbf{y}, A_\textbf{x})|) \\
    & + \log(\gamma_{A_\textbf{z}}\gamma |P(A_\textbf{z}, (A_\textbf{x}, A_\textbf{y}))|) \\
    & - \log (\gamma_{A_\textbf{x}}\gamma|P(A_\textbf{x},(B_\textbf{x}, B_\textbf{y}, B_\textbf{z}))|) \\
    & - \log(\gamma_{A_\textbf{y}}\gamma |P(A_\textbf{y}, (A_\textbf{x},B_\textbf{x}, B_\textbf{y}, B_\textbf{z}))| \\
    & - \log(\gamma_{A_\textbf{z}}\gamma |P(A_\textbf{z}, (A_\textbf{x}, A_\textbf{y}, B_\textbf{x}, B_\textbf{y}, B_\textbf{z}))|)) \\
    & + const. \\
    = & \log \frac{|P(A_\textbf{y}, A_\textbf{x})||P(A_\textbf{z}, (A_\textbf{x}, A_\textbf{y}))| }{|P(A_\textbf{x},(B_{\textbf{x}\textbf{y}\textbf{z}}))| |P(A_\textbf{y}, (A_\textbf{x},B_{\textbf{x}\textbf{y}\textbf{z}}))| |P(A_\textbf{z}, (A_\textbf{x}, A_\textbf{y}, B_{\textbf{x}\textbf{y}\textbf{z}}))|}\\
    & + const.
\end{split}
\end{equation*}
$B_{\textbf{x}\textbf{y}\textbf{z}}$ represent the space constructed by $B_\textbf{x}$, $B_\textbf{y}$ and $B_\textbf{z}$.

Restricted by computational complexity, we approximate it by
\begin{equation*}
\begin{split}
    &\mathbb{I}(\hat{\mathbb{N}}(p_i), \hat{\mathbb{N}}(p_j)) \\
    \approx  & \log \frac{1}{|P(A_\textbf{x},B_\textbf{x})| |P(A_\textbf{y}, B_\textbf{y})| |P(A_\textbf{z},  B_\textbf{z})|} + const.
\end{split}
\end{equation*}
It only considers the relationship of corresponding parts from the weighted sum with respect to parameter gradients. To further simplify, we used a simple formula which computes the cosine similarity of their concatenated gradients as described in main paper.

\section{More Information on Datasets and Evaluation Metrics}

\paragraph{Datasets.}
We report the statistics of the scenes in our evaluation in Tab.~\ref{tab:scannetpp_sta}. 
In addition to the number of images, we calculate the average proportion of overlapping pixels between adjacent images. 

\begin{table*}[!bt]

\caption{
Statistics of the scenes in our evaluation on ScanNet++ and Replica. 
}

\centering
\resizebox{0.96\textwidth}{!}{ 
\begin{tabular}{l|cccccccc|cccc}

\toprule
Datasets & \multicolumn{8}{|c|}{ScanNet++} & \multicolumn{4}{|c}{Replica}\\ 

\midrule
Scene name & 0a7c & 0a18 & 6ee2 & 7b64 & 56a0 & 9460 & a08d & e0ab & office0 & office1 & room0 & room1\\ 

Number of images & 67 & 63 & 75 & 78 & 72 & 79 & 85 & 57 & 60 & 60 & 60 & 60\\
Overlaps (our split) & 0.82 & 0.82 & 0.75 & 0.77 & 0.73 & 0.80 & 0.83 & 0.65 & 0.89 & 0.91 & 0.90 & 0.90\\
Original overlaps & 0.96 & 0.96 & 0.98 & 0.95 & 0.90 & 0.97 & 0.98 & 0.94 & 0.99 & 0.99 & 0.99 & 0.99\\

\bottomrule

\end{tabular}
}

\label{tab:scannetpp_sta}
\end{table*}

\paragraph{Evaluation metrics.}
We use the L2 Chamfer distance and F-score to evaluate the reconstruction results. Both the two metrics are computed on top of the meshes: the ground truth mesh and the reconstructed one. 
The Chamfer distance is computed as: 
\begin{equation}
cd(S_{1}, S_{2}) = \frac{1}{|S_{1}|} \sum_{x\in S_{1}} \min_{y \in S_{2}} ||x-y||_2^2 + \frac{1}{|S_{2}|} \sum_{x\in S_{2}} \min_{y \in S_{1}} ||x-y||_2^2 .
\end{equation}
In this case, $S_1$ and $S_2$ represent the two point sets sampled from the ground truth mesh and the reconstruction, respectively. 

The F-socre is computed as: 
\begin{equation}
fs(S_1, S_2) = 2 \cdot \frac{precision(S_1, S_2)\times recall(S_1, S_2)}{precision(S_1, S_2) + recall(S_1, S_2)} . 
\end{equation}
The values for precision and recall are determined by the proportion of sampled points, where the distance to the nearest point in another mesh is less than 2\% of the scene length. Precision is calculated from the reconstruction to the ground truth, while recall is calculated in the opposite direction.

\begin{figure*}[tb]
\centering
\includegraphics[width=0.75\linewidth]{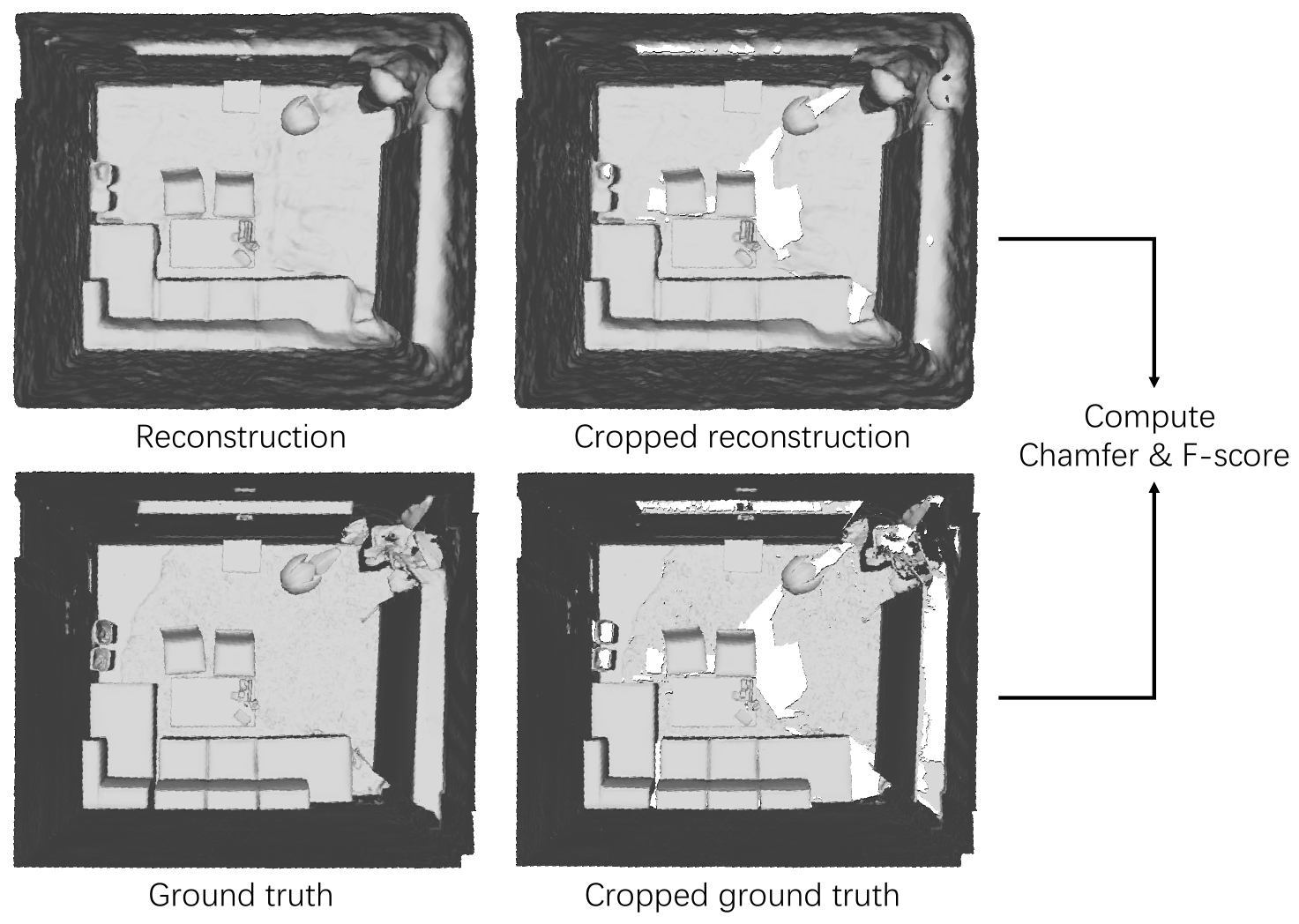}
\caption{
Cropped reconstruction and ground truth with training viewpoints. 
}
\label{fig:cull}
\end{figure*}

We sampled 50,000 points from each original mesh to calculate the metrics.
To ensure a fair evaluation, we remove all parts of the geometry that are not visible from the training views, as shown in Figure~\ref{fig:cull}.

\section{More Implementation Details for Each Baseline} 
The numbers reported in Table~\ref{tab:consume} in the main paper were measured on an NVIDIA H800. 
For all the methods we apply our mutual information shaping to their official codes at GitHub. 

\begin{itemize}
\item NeuS~\cite{wang2021neus}$^+$: https://github.com/Totoro97/NeuS. 
The network is trained by 160k iterations. 

\item VolSDF~\cite{yariv2021volume}$^+$:
https://github.com/lioryariv/volsdf. 
The network is trained by 150k iterations. 

\item GeoNeuS~\cite{fu2022geo}$^+$:
https://github.com/GhiXu/Geo-Neus. 
We use adjacent 8 images (4 before and 4 after) as reference perspectives.
The network is trained by 150k iterations. 

\item I$^2$-SDF~\cite{zhu2023i2sdf}$^+$:
https://github.com/jingsenzhu/i2-sdf. 
We discard the normal and depth supervision.
The network is trained by 150k iterations. 

\item NeuRIS~\cite{wang2022neuris}$^+$:
https://github.com/jiepengwang/NeuRIS. 
The network is trained by 160k iterations. 

\item MonoSDF~\cite{yu2022monosdf}$^+$:
https://github.com/autonomousvision/monosdf. 
We set the decay for the normal and depth loss at 30k iterations. We observed that full use causes the method to degenerate into estimation fusion, rather than reconstruction from posed images.
The network is trained by 100k iterations. 

\item Neuralangelo~\cite{li2023neuralangelo}$^+$:
https://github.com/NVlabs/neuralangelo. 
We set the hash encoding dictionary size to 20 and the feature dimension to 4. This is to ensure VRAM consumption stays at the same level as with other methods.
The network is trained by 150k iterations. 

\end{itemize}

\begin{table}[]
\caption{$\lambda_{M}$ values.}
\centering
\resizebox{0.24\textwidth}{!}{
\begin{tabular}{@{}lc@{}}
    \toprule
    Method & Value\\
    \midrule
    
    NeuS$^+$  & 1.0 \\
    VolSDF$^+$ & 0.3 \\
    
    GeoNeuS$^+$ & 1.0 \\
    $I^2$-SDF$^+$ & 0.3 \\

    NeuRIS$^+$ & 1.0 \\
    MonoSDF$^+$ & 0.5 \\

    Neuralangelo$^+$ & 1.0 \\
    
  \bottomrule
\end{tabular}
}
\label{tab:lambda}
\end{table}
For different baselines, we use different weights $\lambda_M$ to balance the original training and our mutual information shaping. The detailed weights are reported in Table~\ref{tab:lambda}. For all the experiments in the main paper, we set the positive sample threshold with $\beta_S = 0.65$ and $\beta_G=0.99$ for DINO~\cite{caron2021emerging} and normal features~\cite{bae2021estimating}, respectively.

\section{Additional Comparisons}
In the main paper, we exclude GeoNeuS~\cite{fu2022geo}$^+$ and MonoSDF~\cite{yu2022monosdf}$^+$ from the Replica dataset. This is because there are no readily available structure-from-motion models for GeoNeuS, and MonoSDF's training and ablation studies are based on Replica. 
Here, we report their performance in Tab.~\ref{tab:replica_add}. 

\begin{table*}[!tbh]

\caption{
Quantitative results on the Replica dataset.
}

\centering
\resizebox{0.95\textwidth}{!}{ 
\begin{tabular}{cl|cccc|c}

\toprule
 & & office0 & office1 & room0 & room1 & Mean \\ 

\midrule

\multirow{2}{*}{Chamfer (m)$\downarrow$}  
& GeoNeuS~\cite{fu2022geo}$^+$ & 0.0230\textcolor{mygreen}{-0.0206} & 0.0136 \textcolor{mygreen}{-0.0098}  & 0.0396 \textcolor{mygreen}{-0.0354}  & 0.0024 \textcolor{mygreen}{-0.0003}  & 0.0196\textcolor{mygreen}{-0.0165}  \\ 

& MonoSDF~\cite{yu2022monosdf}$^+$ & 0.0028\textcolor{orange}{+0.001} & 0.0047 \textcolor{mygreen}{-0.0006}  & 0.0041 \textcolor{mygreen}{-0.0003}  & 0.0044 \textcolor{mygreen}{-0.0007}  & 0.0040\textcolor{mygreen}{-0.0004}\\ 

\midrule

\multirow{2}{*}{F-score $\uparrow$} 
& GeoNeuS~\cite{fu2022geo}$^+$ & 0.891\textcolor{mygreen}{+0.083} & 0.910 \textcolor{mygreen}{+0.039}  & 0.896 \textcolor{mygreen}{+0.085}  & 0.977 \textcolor{mygreen}{+0.007}  & 0.919\textcolor{mygreen}{+0.054} \\ 

& MonoSDF~\cite{yu2022monosdf}$^+$ & 0.962\textcolor{mygreen}{+0.004} & 0.901 \textcolor{mygreen}{+0.021}  & 0.982 \textcolor{mygreen}{+0.004}  & 0.946 \textcolor{mygreen}{+0.016}  & 0.948\textcolor{mygreen}{+0.012} \\ 

\bottomrule

\end{tabular} 
}

\label{tab:replica_add}
\end{table*}

For GeoNeuS, we utilize COLMAP~\cite{schoenberger2016sfm} to build the structure-from-motion models. We input known camera parameters and perform only triangulation. The table illustrates the advantages of our method to enhance GeoNeuS. Similarly, MonoSDF$^+$ also benefits from our mutual information shaping.

\section{Additional Analyses}
\paragraph{Effectiveness of the semantic features.}
We examine the effectiveness of the semantic feature DINO by replacing it with a semantic segmentation model, SAM~\cite{kirillov2023segment}. The results are shown in Tab.~\ref{tab:ablation_sam}. 
As observed, employing positive-negative pairs with SAM can enhance the baseline performance in some instances (NeuRIS), but it can also hurt performance in other cases (MonoSDF). 
On the contrary, using DINO features consistently enhances both the two baselines. 
Therefore, we apply DINO in our method and report the results with DINO features accordingly in the main paper.

\begin{table*}[!h]
\caption{
Ablation study with image segmentation model - SAM~\cite{kirillov2023segment}.
}

\centering
\resizebox{0.86\textwidth}{!}{ 
\begin{tabular}{l|cccc|cccc}

\toprule

& \multicolumn{4}{|c|}{Chamfer (m)$\downarrow$} 
& \multicolumn{4}{|c}{F-score $\uparrow$} \\
 & 6ee2 & 7b64 & 9460 & Mean & 6ee2 & 7b64 & 9460 & Mean \\ 

\midrule

NeuRIS~\cite{wang2022neuris} 
& \textbf{0.029} & 0.070 & 0.405 & 0.168 & 0.65 & 0.69 & 0.24 & 0.53 \\ 
NeuRIS$^+$ (SAM) & \underline{0.038} & 0.042 & \underline{0.337} & \underline{0.139} & \textbf{0.69} & 0.73 & \underline{0.30} & \underline{0.57}  \\ 
NeuRIS$^+$ (SAM+normal) & 0.053 & \textbf{0.030} & 0.412 & 0.165 & \underline{0.68} & \textbf{0.77} & 0.22 & 0.56   \\ 
NeuRIS$^+$ (Full) & 0.044 & \underline{0.033} & \textbf{0.198} & \textbf{0.092} & 0.66 & \underline{0.76} & \textbf{0.41} & \textbf{0.61} \\ 

\midrule

MonoSDF~\cite{yu2022monosdf} & \underline{0.020} & \textbf{0.016} & 0.046 & \underline{0.028} & \underline{0.81} & 0.79 & \underline{0.88} & \underline{0.83} \\ 
MonoSDF$^+$ (SAM) & 0.527 & \underline{0.018} & 0.035 & 0.193 & 0.63 & 0.77 & 0.86  & 0.75 \\ 
MonoSDF$^+$ (SAM+normal) & 0.062 & 0.018 & \underline{0.029} & 0.037 & 0.77 & \underline{0.80} & 0.88  & 0.82 \\ 
MonoSDF$^+$ (Full) & \textbf{0.014} & 0.020 & \textbf{0.023} & \textbf{0.019} & \textbf{0.85} & \textbf{0.85} & \textbf{0.93} & \textbf{0.87} \\ 

\bottomrule

\end{tabular} 
}

\label{tab:ablation_sam}
\end{table*}

\begin{table} [htb!]

\caption{
Ablation study for the first-order method with correlated normals. 
}

\centering
\resizebox{0.60\textwidth}{!}{ 
\begin{tabular}{cl|ccc|c}

\toprule
 & & 6ee2 & 7b64 & 9460 & Mean \\ 

\midrule

\multirow{4}{*}{\rotatebox{90}{Chamfer$\downarrow$}}  

& NeuRIS$^+$ (FO) & 0.586 & 0.490 & 0.603 & 0.560 \\ 
& NeuRIS$^+$ (Full) & \textbf{0.044} & \textbf{0.033} & \textbf{0.198} & \textbf{0.092} \\ 

\cline{2-6}

& MonoSDF$^+$ (FO) & 1.207& 1.024 & 1.163&1.132\\ 
& MonoSDF$^+$ (Full) & \textbf{0.014} & \textbf{0.020} & \textbf{0.023} & \textbf{0.019} \\ 

\midrule

\multirow{4}{*}{\rotatebox{90}{F-score$\uparrow$}} 

& NeuRIS$^+$ (FO) & 0.43 & 0.34 & 0.21 & 0.33\\ 
& NeuRIS$^+$ (Full) & \textbf{0.66} & \textbf{0.76} & \textbf{0.41} & \textbf{0.61} \\ 

\cline{2-6}

& MonoSDF$^+$ (FO) & 0.59& 0.49& 0.57& 0.55\\ 
& MonoSDF$^+$ (Full) & \textbf{0.85} & \textbf{0.85} & \textbf{0.93} & \textbf{0.87} \\ 

\bottomrule

\end{tabular}

}

\label{tab:first-order}

\end{table}

\paragraph{Performance of first-order method.}
In Tab.~\ref{tab:first-order}, 
we report the results of directly aligning the normal directions (denoted by FO)  among positive pairs (i.e., correlated surfaces). It is implemented by replacing $L_M$ with 

\begin{equation}
    L_M^\prime = -\log(\sum \exp(||\cos(\mathbb{N}_i, \mathbb{N}_{i+}||)) ,
\end{equation}

which is similar to Eq.~\ref{equ:infonce} but removes the calculation of second-order and the part of negative pairs. 
We notice performance drops, and the results appear over-smoothed. This is mainly because a) the positive pairs have similar but not identical directions, and b) the information from negative pairs is not utilized. 

\begin{table} [htb!]

\caption{
Quantitative results on the DTU dataset. 
}

\centering
\resizebox{0.55\textwidth}{!}{ 
\begin{tabular}{cl|ccc|c}

\toprule
 & & 24 & 37 & 40 & Mean \\ 

\midrule

\multirow{4}{*}{\rotatebox{90}{Chamfer$\downarrow$}}

& NeuRIS~\cite{wang2022neuris} & \textbf{0.980} & 3.674 &  0.865 & 1.840\\ 

& NeuRIS$^+$ & 1.023 & \textbf{3.341} & \textbf{0.663} & \textbf{1.676} \\ 

\cline{2-6}

& MonoSDF~\cite{yu2022monosdf} & 0.876 &  \textbf{1.773} & 0.657 & 1.102 \\ 

& MonoSDF$^+$ & \textbf{0.837} & 1.816 & \textbf{0.626} & \textbf{1.093} \\ 

\bottomrule

\end{tabular}

}

\label{tab:dtu}

\end{table}

\paragraph{More discussions on limitation. }
While our method does not rely on Manhattan world or near-planar assumptions, we have found that its effectiveness on object-level scenes is reduced. 
In Tab~\ref{tab:dtu}, we present the quantitative results on object-level inward-facing scenes from the DTU dataset.
The experiments are carried out in the first three scenes with two baselines, using the same hyperparameter settings as those used in the indoor scenes. 
In these scenes, we note that DINO features or monocular normals sometimes produce inconsistent pairs, which could affect the effectiveness of our mutual information shaping.

%
%

\bibliographystyle{splncs04}

\end{document}